%% file: IEEE-conference-template-062824.tex
\documentclass[conference]{IEEEtran}
\IEEEoverridecommandlockouts

\usepackage{cite}
\usepackage{amsmath,amssymb,amsfonts}
\usepackage{algorithm}
\usepackage{algorithmic}
\usepackage{caption}
\usepackage{subcaption}
\usepackage{tikz}
\usepackage{graphicx}
\usepackage{textcomp}
\usepackage{xcolor}
\usepackage{multirow}
\usepackage[table,xcdraw]{xcolor}
\usepackage{hyperref}
\usepackage{amsmath}
\usepackage{hhline}

\def\BibTeX{{\rm B\kern-.05em{\sc i\kern-.025em b}\kern-.08em
    T\kern-.1667em\lower.7ex\hbox{E}\kern-.125emX}}

\usepackage{tablefootnote}
\PassOptionsToPackage{table}{xcolor}

\begin{document}

\title{Efficient LLMs with AMP: \underline{A}ttention Heads and \underline{M}LP \underline{P}runing}

%
%

\author{%
  \IEEEauthorblockN{%
    Leandro Giusti Mugnaini\IEEEauthorrefmark{1}\textsuperscript{\textsection},
    Bruno Lopes Yamamoto\IEEEauthorrefmark{1}\textsuperscript{\textsection},
    Lucas Lauton de Alcantara\IEEEauthorrefmark{1},
    Victor Zacarias\IEEEauthorrefmark{1}, \\
    Edson Bollis\IEEEauthorrefmark{2},
    Lucas Pellicer\IEEEauthorrefmark{2},
    Anna Reali\IEEEauthorrefmark{1} and
    Artur Jordao\IEEEauthorrefmark{1}%
  }%
  \IEEEauthorblockA{\IEEEauthorrefmark{1} Escola Politécnica de Engenharia, Universidade de São Paulo, São Paulo, Brazil}%
  \IEEEauthorblockA{\IEEEauthorrefmark{2} Instituto de Ciência e Tecnologia Itaú (ICTi), São Paulo, Brazil}%
}

\maketitle

\begingroup
\renewcommand\thefootnote{\textsection}
\footnotetext{Equal contribution.}
\endgroup

\input{Sections/abstract}
\input{Sections/Introduction}
\input{Sections/RelatedWork}

%
%
\input{Sections/AMP_new_notation}

%
\input{Sections/Experiments}

\input{Sections/Discussion}

\input{Sections/Acknowledgment}

\bibliographystyle{IEEEtran}
\bibliography{refs}

\end{document}

%% file: Sections/Abstract.tex
\begin{abstract}

Deep learning drives a new wave in computing systems and triggers the automation of increasingly complex problems. In particular, Large Language Models (LLMs) have significantly advanced cognitive tasks, often matching or even surpassing human-level performance. However, their extensive parameters result in high computational costs and slow inference, posing challenges for deployment in resource-limited settings. Among the strategies to overcome the aforementioned challenges, pruning emerges as a successful mechanism since it reduces model size while maintaining predictive ability. In this paper, we introduce AMP: \underline{A}ttention Heads and \underline{M}LP \underline{P}runing, a novel structured pruning method that efficiently compresses LLMs by removing less critical structures within Multi-Head Attention (MHA) and Multilayer Perceptron (MLP). By projecting the input data onto weights, AMP assesses structural importance and overcomes the limitations of existing techniques, which often fall short in flexibility or efficiency. In particular, AMP surpasses the current state-of-the-art on commonsense reasoning tasks by up to 1.49 percentage points, achieving a 30\% pruning ratio with minimal impact on zero-shot task performance. Moreover, AMP also improves inference speeds, making it well-suited for deployment in resource-constrained environments. We confirm the flexibility of AMP on different families of LLMs, including LLaMA and Phi.

\end{abstract}


%% file: Sections/Introduction.tex
\section{Introduction}\label{sec:introduction}




Within the evolving landscape of Artificial Intelligence, Large Language Models (LLMs) stand out as a pivotal force, propelling Natural Language Processing towards unprecedented stages -- often matching or even surpassing human-level performance in many language benchmarks~\cite{openai2024gpt4,zhou:2023,Brown:2020}.
However, this performance often comes at the cost of larger model sizes, with models such as DeepSeek-V3~\cite{deepseekai2024} reaching the mark of 671 billion parameters.

This substantial size introduces significant challenges for the deployment of LLMs in low-resource and time-critical applications due to high computational costs and slow inference~\cite{Sheng:2023}.
To overcome these challenges, some techniques such as pruning~\cite{Zhu:2024} and quantization~\cite{frantar2023optq,Renren:2024} aim to reduce the model size, thereby lowering computational overhead.

Recent studies confirm pruning as a promising solution to compress models as it maintains predictive ability and is often hardware-agnostic~\cite{Ma:2023,Sun:2024,Gao:2024disp,Ashkboos:2024,Ouderaa:2024}.
%
Within the field of LLMs, pruning techniques fall into three main categories: structured, semi-structured and unstructured pruning.

Structured pruning removes entire components -- such as attention heads or layers -- while preserving the overall network structure without introducing sparsity (i.e., without zeroing out a significant portion of the model’s parameters)~\cite{Zhu:2024}. However, the removal of larger and potentially more critical components may result in performance degradation, typically requiring Parameter-Efficient Fine-Tuning (PEFT) techniques for performance recovery~\cite{Han:2024}. Due to the removal of complete components, structured pruning usually achieves inference acceleration and memory reduction without the need for specialized hardware or software~\cite{Kim:2024}.

Semi-structured (a.k.a. structured $N:M$) pruning promotes model sparsity by removing groups of consecutive parameters following a pruning mask~\cite{Frantar:2023}. Specifically, structured $N:M$ sparsity requires that at least $N$ out of every $M$ consecutive weights be non-zero~\cite{zhou2021learning,Mishra:2021}. While this is a promising technique, it requires specialized hardware to achieve practical speedup, making it less suitable for deployment on consumer-grade GPUs~\cite{zhou2021learning}.

Finally, unstructured pruning removes  weights without considering any pattern, increasing model sparsity by zeroing out parameters~\cite{Frantar:2023,Sun:2024}. However, the resulting irregular sparse matrices often fail to achieve efficiency comparable to dense matrices on general-purpose hardware~\cite{Mishra:2021,Gao:2024disp}. To improve efficiency, the resulting architecture requires specific software or hardware optimizations, often resulting in lower inference speedups when compared to other forms of pruning~\cite{Mishra:2021}.



Regardless of the category, pruning relies on importance metrics to guide its decisions. For example, magnitude-based methods use the absolute value of weights to determine less important structures~\cite{Frantar:2023,Kim:2024}.
In this context, a data-driven approach takes weights and input data into account on component evaluation~\cite{Sun:2024}. In contrast, loss-based metrics assess structure importance by measuring their impact on the loss function.
%
Successful methods in this sphere typically employ Taylor expansion~\cite{Ma:2023,Kim:2024,wang:2019eigendamage}.
 Different techniques also encompass regularization into their methodology~\cite{Wang:2021, Zhang:2022}. Finally, some approaches consider the similarity between weights or representations, employing metrics such as cosine similarity to determine the least important structures~\cite{gromov:2024}.

Despite the effectiveness of the aforementioned strategies, existing pruning methods often face trade-offs involving performance, latency, or the requirement for specialized hardware. Hence, there remains a need for simpler, hardware-agnostic techniques that effectively compress LLMs with minimal performance loss while achieving tangible inference speedups.
%
In this work, we introduce \underline{A}ttention Heads and \underline{M}LP \underline{P}runing (AMP), a pruning technique that addresses gaps in existing methods. Our criterion removes attention heads and MLP neurons across layers to achieve target compression rates. AMP uses the magnitude of activations -- rather than solely the weights -- to evaluate the importance of structures within each layer, ensuring uniform pruning throughout all layers. Among our contributions, we highlight the following:
\begin{itemize}
	\item We propose AMP, an efficient and effective structured pruning method that surpasses the state-of-the-art techniques of LLM pruning, providing practical inference speedups without the need for specialized hardware. Notably, our method achieves up to $1.25\times$ inference speedup while surpassing existing structured pruning techniques by up to 1.49 percentage points in average task accuracy.
    \item In contrast to existing approaches, AMP identifies unimportant structures within minutes and requires a negligible fine-tuning process for performance recovery, making it both practical and resource-efficient. 
    \item We reveal that projecting input data onto weights successfully discerns critical from non-critical components. We confirm the effectiveness of our strategy with a coherence experiment, an essential validation we believe should become standard in pruning research.
    \item Our method is applicable to different LLM families, requiring minimal modifications.
    Additionally, AMP is aligned with Green AI principles when deploying LLMs at scale, leading to direct cost savings and reduced CO$_2$ emissions.
 
\end{itemize}

Code and models are available at: \url{https://github.com/c2d-usp/Efficient-LLMs-with-AMP}.

%% file: Sections/RelatedWork.tex
\section{Related Work}\label{sec:related}

We contrast our method with the three categories of pruning~\cite{Wan:2023,Zhu:2024}: structured, semi-structured and unstructured.

\noindent \textbf{Structured Pruning.} LLM-Pruner~\cite{Ma:2023} locates discrete neuron clusters in LLMs based on their internal dependencies, measures the importance of each one based on loss changes, and prunes those with lower importance. Sheared LLaMA~\cite{Xia:2024 } compresses LLMs into compact architectures and continues pre-training with dynamic batch loading, using loss values to guide its pruning and training processes.

Unlike previous works, Shortened LLaMA~\cite{Kim:2024} eliminates entire Transformer layers using Taylor expansion (based on differences in training loss) or a perplexity-driven criterion (based on perplexity changes). 
SliceGPT~\cite{Ashkboos:2024} prunes models by removing rows and columns according to Principal Component Analysis, achieving strong performance. However, it requires the addition of many extra parameters to maintain compatibility with internal model dimensions.

Similarly to SliceGPT, DISP-LLM~\cite{Gao:2024disp} eliminates modules by removing rows and columns via selection matrices and employs ReinMax to optimize layer width. In contrast, FLAP~\cite{An:2024} multiplies the variance of a calibration dataset by the squared weights norm to score channels, standardizes these scores, removes channels accordingly, and adjusts bias. As a result, FLAP not only depends heavily on the calibration dataset but also introduces bias into the model. According to Xia et al.~\cite{Xia:2024}, the varying layer widths introduced by these methods lead to additional training and inference overhead.

Alternatively, PruneNet~\cite{YOPO:2025} employs a policy learning-based compression for LLMs. Essentially, policy gradient trains a learner model penalizing it with Kolmogorov-Smirnov distance between singular value distributions for pruned and unpruned matrices. Despite an innovative approach, the study leaves pruning mixed attention heads and MLPs unaddressed, a field we explore through AMP. LLM Surgeon~\cite{Ouderaa:2024} provides a comprehensive framework for unstructured, semi-structured, and structured pruning, employing Kronecker-factored curvature approximations of the target loss landscape for comprehensive pruning. While the method achieves competitive results, it requires significant computational resources. 

In contrast to the existing approaches we mentioned above, our AMP strategy prunes attention heads and MLP neurons by using activation magnitudes from a small number of forward passes, rather than relying on computationally expensive loss-based or projection-driven heuristics. Our effective design avoids rigid constraints, such as  pruning only the Multilayer Perceptron. By applying uniform pruning across layers, AMP also prevents the creation of layers with varying widths, thereby avoiding additional overhead.

\noindent
\textbf{Semi-Structured Pruning.} COPAL~\cite{Malla:2024} prunes weights in Large Language Models by analyzing directional derivatives of the loss function across datasets. 
Although effective, it requires numerous differential operations to determine importance, thereby increasing its computational cost. Wanda~\cite{Sun:2024} identifies sparse sub-networks in LLMs by ranking each weight using the product of its absolute value and its squared input feature norm. However, as Gao et al.~\cite{Gao:2024disp} argue, Wanda excels only as an unstructured pruning technique, a form of sparsity that does not lead to inference speedups in the resulting model. Unlike these methods, AMP does not require specialized hardware to achieve practical speedups. Furthermore, while existing techniques are constrained to fixed sparsity patterns such as 2:4 or 4:8, limiting their compression rates, AMP offers greater flexibility in achieving desired compression levels without being bound by predefined sparsity structures.

\noindent \textbf{Unstructured Pruning.} 
SparseGPT~\cite{Frantar:2023} formulates pruning as a sparse regression solved via an efficient approximate mechanism, achieving up to 60\% sparsity while preserving model performance without traditional retraining. However, it performs less accurately on small to medium-sized models compared to larger networks. Plug-and-Play~\cite{Zhang:2024plug} introduces a one-shot post-training pruning combining Relative Importance and Activation (RIA) with Channel Permutation. Despite avoiding retraining, it exhibits sensitivity to calibration data and increases complexity, particularly for larger models. Moreover, these methods often fail to provide practical inference speedups, as modern GPUs accelerate inference for semi-structured patterns (e.g., 2:4)~\cite{zhou2021learning,Mishra:2021}. On the other hand, AMP follows a structured approach and achieves tangible inference speedups using a lightweight, efficient procedure.




%% file: Sections/AMP_new_notation.tex
\section{Proposed AMP method}
\noindent
Residual connections link the input of different components of Large Language Models to its output, creating a continuous flow of information through the network. This path of information is commonly referred to as the residual stream~\cite{Elhage:2021}. 


Building on this concept, our method measures the importance of each component based on its addition to the residual stream, making it possible to prune components with lower contribution. In particular, we prune  attention heads from Multi-Head Attention (MHA) and neurons from Multilayer Perceptron (MLP) within each layer of the model. This approach provides a more faithful representation of how the network internally processes information than simple weight-only or activation-only heuristics.

The idea behind our method is simple; however, due to the sophisticated architecture of Transformers, we start by defining notations and explaining the two main components involving our AMP criterion: Multi-Head Attention (MHA) and Multilayer Perceptron (MLP). Although the description and explanation focus on the LLaMA-2 7B model, the method is applicable to other Transformer-based LLMs with similar architecture, as we shall see in the experiments.



\noindent
\subsection{Multi-Head Attention}
Each Multi-Head Attention (MHA) mechanism involves several attention heads~\cite{Vaswani:2017}. As these are independent units, they process the input data in parallel. MHA then adds the combined output to the residual stream.

In each attention component of the Transformer architecture,  MHA comprises exactly $N$ attention heads. Consider $S$ the sequence length and $d_{model}$ the hidden size dimension. Given an input \(\mathbf{X} \in \mathbb{R}^{\text{S} \times d_{\text{model}}}\), each attention head $n$ within MHA performs projections into the Key ($\mathbf{K}$), Query ($\mathbf{Q})$ and Value ($\mathbf{V}$) matrices as follows:


\begin{equation}
\mathbf{K}_n = \mathbf{X} \mathbf{W}^K_n, \mathbf{Q}_n = \mathbf{X} \mathbf{W}^Q_n, \mathbf{V}_n = \mathbf{X} \mathbf{W}^V_n,
\end{equation}

\noindent
where $\mathbf{W}^K_n$, $\mathbf{W}^Q_n$ and $\mathbf{W}^V_n$ are learnable projection matrices belonging to $\mathbb{R}^{d_{model} \times d_k}$ and $d_k = \frac{d_{model}}{N}$ is the dimensionality allocated to each attention head.



Building on these projections, the Scaled Dot-Product Attention (SDPA), a fundamental mechanism in Transformer, computes each attention head in terms of

\begin{equation}
\text{Attention}(\mathbf{K}, \mathbf{Q}, \mathbf{V}) = \text{Softmax}\left(\frac{\mathbf{QK}^T}{\sqrt{d_k}}\right)\mathbf{V}.
\end{equation}

Then, for a given layer, we express the \(n\)-th attention head as:

\begin{equation}
\mathbf{h}_n = \text{Attention}(\mathbf{K}_n, \mathbf{Q}_n, \mathbf{V}_n).
\end{equation}



MHA concatenates all \(N\) attention heads and multiplies them by the output matrix \(\mathbf{W}_O \in \mathbb{R}^{d_{model} \times d_{model}}\). Specifically,

\begin{equation}\label{eq::MHA}
\text{MHA}(\mathbf{X}) = \text{Concat}(\mathbf{h}_1, \mathbf{h}_2, \dots, \mathbf{h}_N)\mathbf{W}_O.
\end{equation}

The LLaMA-2 7B model has a single MHA module in each layer~\cite{Touvron2:2023}, which integrates 32 attention heads (i.e., \textit{N}=32).

\noindent
\subsection{Multilayer Perceptron}
Each LLaMA layer contains a Multilayer Perceptron (MLP) component that uses the SwiGLU mechanism. This component consists of three linear projections: Up, Gate, and Down Projections. Let $x\in \mathbb{R}^{S \times d_{model}}$ be the input tensor. 
The SwiGLU MLP is then the following: 



\begin{equation}\label{eq::MLP_SiLU}
\text{MLP}(x) = \left(\text{SiLU}(x\mathbf{W}_{Gate}) \odot (x\mathbf{W}_{Up}\right)) \mathbf{W}_{Down},
\end{equation}

\noindent
where \( \mathbf{W}_{Up} \) \( \in \mathbb{R}^{d_{model} \times d_i} \), \( \mathbf{W}_{Gate} \) \( \in \mathbb{R}^{d_{model} \times d_i} \), \text{ and } \( \mathbf{W}_{Down} \) \( \in \mathbb{R}^{d_i \times d_{model}}\). $d_i$ is the intermediate size of MLP and $\odot$ represents an element-wise multiplication.


The role of \(\mathbf{W}_{Up}\) and \(\mathbf{W}_{Gate}\) is to project the input $x$ into a higher-dimensional space ($\mathbb{R}^{d_i}$) that enhances representational capacity. 
The projections are then combined element-wise before being projected back to $\mathbb{R}^{d_{model}}$ by \( \mathbf{W}_{Down} \).

\subsection{Proposed AMP method}

 We define our AMP importance criterion as a combination of two independent contributions: the MHA component and the MLP component.

\noindent
\textbf{AMP -- MHA Component.}
The objective of the MHA component of AMP is to define an importance score for each attention head in a layer, determining which ones are less important for the overall performance of the model, and hence targets for pruning.

Due to the fact that $\mathbf{W}_O \in \mathbb{R}^{({d_k} \cdot {N}) \times d_{model}}$, the MHA output will always have size $d_{model}$, regardless of the number of heads $N$. Therefore, removing an entire attention head maintains compatibility with the dimensions of the residual stream and, consequently, with subsequent layers. To completely remove a head, we only need to prune the associated weights from the \textbf{K}, \textbf{Q}, and \textbf{V} matrices. Since it reduces the dimension of the concatenated MHA output, we prune the corresponding part of \(\mathbf{W}_O\) along the appropriate axis.



A naive pruning approach might extend the classical $\ell_p$-norm method -- common in unstructured pruning~\cite{Cheng:2024} -- to structured pruning by computing norms over all parameters within a single attention head. However, this strategy fails to account for the nonlinear interactions between weights and inputs. Hence, an effective metric should assess the impact of each head on the final output of the MHA mechanism.

Simply measuring the output of each head ($h_1, \ldots, h_N$) is equally insufficient, as it neglects the subsequent \(\mathbf{W}_O\) projection. Additionally, evaluating only the final MHA output does not allow the scoring of each head, since the individual contributions are already combined by \(\mathbf{W}_O\) at this stage (see Equation~\ref{eq::MHA}).

To address the previous limitations, the MHA component of our AMP criterion assesses the contribution of each head while incorporating the effect of the \(\mathbf{W}_O\) projection. To accomplish this, we rewrite the final \(\mathbf{W}_O\) linear projection and the concatenated output of the heads \(\mathbf{H}\) as block matrices: 


\[
\mathbf{W}_O = \begin{bmatrix} \mathbf{W}_1 \\ \mathbf{W}_2 \\ \vdots \\ \mathbf{W}_N \end{bmatrix}, \quad \mathbf{W}_n \in \mathbb{R}^{d_{\text{head}} \times d_{\text{model}}}, \quad n \in \{1, 2, \ldots, N\},
\]
\[
\mathbf{H} = \begin{bmatrix} \mathbf{h}_1 & \mathbf{h}_2 & \cdots & \mathbf{h}_N \end{bmatrix}, \quad \mathbf{h}_n \in \mathbb{R}^{S \times d_{\text{head}}}.
\]

Then, the output of MHA becomes
\begin{equation}
\mathbf{O} = \mathbf{H} \mathbf{W}_o = 
\begin{bmatrix} \mathbf{h}_1 & \mathbf{h}_2 & \cdots & \mathbf{h}_N \end{bmatrix} 
\begin{bmatrix} \mathbf{W}_1 \\ \mathbf{W}_2 \\ \vdots \\ \mathbf{W}_N \end{bmatrix},
\end{equation}
and in a summation form
\begin{equation}
\mathbf{O} = \mathbf{h}_1\mathbf{W}_1 + \mathbf{h}_2\mathbf{W}_2 + \dots + \mathbf{h}_N\mathbf{W}_N = \sum_{n=1}^{N} \mathbf{h}_n \mathbf{W}_n.
\label{eq:sum}
\end{equation}

Equation \ref{eq:sum} shows that it is possible to write the output of MHA as a sum of terms $\mathbf{h}_n\mathbf{W}_n$, where each term depends only on the output of its respective attention head. This is effectively representing the output of  MHA as a sum of the contributions of each attention head.


Finally, we define the importance \( \mathbf{I}_n \) of each head as the $\ell_1$ norm of $\mathbf{h}_n\mathbf{W}_n$
\begin{equation}
\mathbf{I}_n = \left\lVert \mathbf{h}_n \mathbf{W}_n \right\rVert_1.
\label{eq:mha_imp}
\end{equation}

Our importance criterion given by Equation~\ref{eq:mha_imp} measures the magnitude of the contribution of each attention head in a layer to the final output of  MHA and, hence, to the residual stream, by projecting the data onto the weights, differing from the common pruning strategy of simply applying the $\ell_p$-norm. This distinction highlights the unique approach of measuring the importance of attention heads, setting it apart from traditional pruning techniques.

\noindent
\textbf{AMP -- MLP Component.} 
We propose pruning neurons from the up and gate projections in the SwiGLU MLP (see Equation \ref{eq::MLP_SiLU}); thereby reducing the MLP's intermediate size ($d_i$) while maintaining the hidden dimension ($d$). The SwiGLU mechanism employs element-wise multiplication between the up and gate outputs, so the neurons must be pruned in pairs from both projections.
To identify which pairs to prune, the MLP component of AMP accounts for both weight parameters and input magnitudes. In contrast to simple $\ell_1$ or $\ell_2$ norms of the weights -- whose scores could be sometimes misleading as they ignore the scale of the input~\cite{Frantar:2023} -- the MLP component measures how strongly each pair of up-gate neurons contributes to the overall activation and, consequently, to the residual stream. Formally, for a set of input tokens $x$ and the $m$-th pair of neurons, we compute the $l_1$-norm of element-wise product of the outputs from the gate and up projections, and then average over all tokens in the sample (i.e., a sentence). This is effectively the $l_1$-norm of the input of the Down Projection, as we formalize in terms of:  

\begin{equation}
\mathbf{I}_m = \frac{1}{S} \sum_{s=1}^{S} \left| \text{SiLU}\left((x_s \mathbf{W}_{Gate})_m\right) \cdot (x_s \mathbf{W}_{Up})_m \right|,
\label{eq:mlp_imp}
\end{equation}

\noindent
where ${x_s}$ is the MLP input of the token $s$ and $\text{SiLU}(x_s \mathbf{W}_{Gate})_m$ represents the $m$-th element of the activation vector $\text{SiLU}(x_s \mathbf{W}_{Gate})$. Similarly, $(x_s \mathbf{W}_{Up})_m$ indicates the $m$-th element of the projection vector $x_s \mathbf{W}_{Up}$.





\begin{algorithm}[!b]
	\small
	\footnotesize
	\caption{Pruning with our AMP method}
	\label{alg::pruning}
	\begin{algorithmic}[1]
		\item[] \textbf{Input:} Large Language Model $\mathcal{F}$,  Compression rate $c$, Training samples $\textbf{X}$ and the respective labels $\textbf{Y}$, Subset of samples  $\textbf{X}_{sub} \subset \textbf{X}$
		\item[] \textbf{Output:} $\mathcal{F}^{'}$ (Pruned version of $\mathcal{F}$)\\
    
        \STATE $AMP_{\text{MHA}} \gets AMP(\mathcal{F}, \textbf{X}_{sub})$ $\triangleright$ Extract mean activation values for attention heads using Eq.~\ref{eq:mha_imp}

        \STATE $AMP_{\text{MLP}} \gets AMP(\mathcal{F}, \textbf{X}_{sub})$ $\triangleright$ Extract mean activation values for MLP neurons using Eq.~\ref{eq:mlp_imp}

        \FOR{each layer $\ell \in L$} 
            \STATE $AMP_{\text{MHA}_\ell} \gets sorted(AMP_{\text{MHA}}[:, \ell])$ $\triangleright$ Sorts attention heads of layer $\ell$ by importance
            \STATE $AMP_{\text{MLP}_\ell} \gets sorted(AMP_{\text{MLP}}[:, \ell])$ $\triangleright$  Sorts MLP neurons of layer $\ell$ by importance
            \STATE$\mathcal{F}^{'}_{\ell} \leftarrow \mathcal{F}_{\ell} \setminus (AMP_{\text{MHA}_\ell}, c) $ $\triangleright$ Remove unimportant heads from $\ell$ based on $AMP_{\text{MHA}_\ell}$ and compression rate $c$
            \STATE$\mathcal{F}^{'}_{\ell} \leftarrow \mathcal{F}^{'}_{\ell} \setminus (AMP_{\text{MLP}_\ell}, c) $ $\triangleright$ Remove unimportant MLP neurons from  $\ell$ based on $AMP_{\text{MLP}_\ell}$ and compression rate $c$

        \ENDFOR

		\STATE Update $\mathcal{F}^{'}$ via fine-tuning on  $\textbf{X}$ and $\textbf{Y}$

	\end{algorithmic}
\end{algorithm}

\noindent
\textbf{Pruning Process with AMP.}
Given a Large Language Model $\mathcal{F}$ composed of a layer set $L$ and $P$ parameters, our goal is to remove attention heads and MLP neurons from each layer $\ell \in L$ to derive a compressed network $\mathcal{F}'$ with $Q$ parameters, where $Q \ll P$. We define the pruning ratio $c$ as $(1 - Q / P) \times 100$ to quantify the percentage reduction in parameters from the original model to the pruned version.

Building upon the previous formalism, the pruning with our AMP is the following.
Let $\mathbf{X}$ denote training samples, such as sentences, and $\mathbf{Y}$ the respective labels. Following previous studies~\cite{Ma:2023, Zhang:2024plug, Kim:2024, Sun:2024}, we consider random samples from the training set $\mathbf{X}$, denoting it as $\mathbf{X}_{sub}$. In particular, these works confirm that a small calibration set is sufficient to capture key input statistics; therefore, to make a fair comparison and reduce the computational costs we follow the same practice.

$AMP(\cdot, \mathbf{X}_{sub})$ is the AMP method that extracts the mean feature representations (activations) from MLP neurons and attention heads, according to Equations \ref{eq:mha_imp} and \ref{eq:mlp_imp}, for each layer $\ell$ using the samples in $\mathbf{X}_{sub}$.

We determine the final importance of each component using the activation values. To achieve the desired compression rate $c$, we prune $c\%$ of attention heads and $c\%$ of MLP neurons with the lowest importance uniformly across all layers, preserving the layer-modular architecture of the Transformer. Algorithm \ref{alg::pruning} summarizes the pruning process with AMP.

%% file: Sections/Experiments.tex
\section{Experiments}\label{sec:experiments}

\subsection{Experimental Settings}

\noindent
\textbf{Large Language Models.} In order to compare our method against existing pruning techniques, we consider four open-source LLMs commonly used in pruning research: LLaMA 7B~\cite{Touvron:2023}, LLaMA-2 7B~\cite{Touvron2:2023}, Phi-1.5 and Phi-2~\cite{Li:2023}. 

\noindent
\textbf{Evaluation and Datasets.} We evaluate the mean performance of each model at different pruning ratios. For a fair comparison, we opt for benchmarks of previous works~\cite{Ma:2023,Kim:2024,Gao:2024disp,Ashkboos:2024,Ouderaa:2024}. We use the EleutherAI LM Harness framework~\cite{Gao:2024} to perform zero-shot task classification on common sense reasoning datasets: WinoGrande~\cite{Sakaguchi:2019}, HellaSwag~\cite{Zellers:2019}, ARC-e / ARC-c~\cite{Clark:2018} and PIQA~\cite{Bisk:2020}. Following Ouderaa et al.~\cite{Ouderaa:2024}, we report the normalized accuracy for all benchmarks, except for WinoGrande, where we report accuracy. Additionally, we perform a perplexity (PPL) analysis on WikiText2~\cite{Merity:2016}, using the procedures by Ashkboos et al.~\cite{Ashkboos:2024}.

\noindent
\textbf{Inference Speedup Evaluation.} To evaluate inference speedup, we adopt the definition of latency from Sheng et al.~\cite{Sheng:2023}. For a batch size $J$ and an output sequence length $K$, latency $T$ is defined as the time required to generate $JK$ output tokens based on the input prompts. We measure speedup as the relative difference in latency between the pruned model and its original, unpruned, counterpart. Following Kim et al.~\cite{Kim:2024}, we compute latency using 12 input tokens, 128 output tokens and a batch size of 1 on a single GPU. We report the average results over 20 runs, excluding the initial 10 warm-up batches.

\noindent
\textbf{Implementation Details.} In order to compute the activations necessary for the method, we use 50 random samples from the cleaned version of the Alpaca dataset~\cite{alpaca}. This choice aligns with prior studies that consider a small calibration set (typically ranging from 10 to 128 samples) to conduct the pruning process~\cite{Ma:2023,Zhang:2024plug,Kim:2024,Sun:2024}. Following Ma et al.~\cite{Ma:2023}, we employ a post-training process using low-rank adaptation, LoRA~\cite{Hu:2022}, to recover the model performance. We set the LoRA rank to 8, learning rate to 3e-4, batch size to 1 and sequence length to 512, using an AdamW optimizer.
We use the Alpaca dataset and fine-tune the model for 2 epochs on a single consumer-grade GPU (NVIDIA RTX 3090). It is worth mentioning that this setup follows previous work~\cite{Ma:2023} and ensures the benefits of our method do not come from hyperparameter customizations.

\noindent
\subsection{Zero-shot Performance}
\label{exp:main_experiment}

We start our experiments by comparing the AMP criterion with top-performing structured pruning techniques. For a fair comparison, we report the results of each method according to the original paper and prune the models using the same compression rates. Table \ref{tab:state_of_the_art} presents the results.

\setlength{\doublerulesep}{1pt}

\begin{table*}[!tb]
\centering
\caption{Comparison of state-of-the-art structured pruning methods across different LLMs. }
\begin{tabular}{c|c|ccccc|c}
\hline
Pruning Ratio          & Method                                            & WinoGrande                             & HellaSwag                              & ARC-e                                  & ARC-c                                  & PIQA                                   & Avg                                    \\ \hhline{=|=|=====|=}
0\%                    & LLaMA 7B                                        & 69.85                                  & 76.21                                  & 72.81                                  & 44.71                                  & 79.16                                  & 68.55                                  \\ \hline
                       & \cellcolor[HTML]{EFEFEF}LLM Pruner~\cite{Ma:2023} (NeurIPS, 2023)                                        & \cellcolor[HTML]{EFEFEF}61.33                                  & \cellcolor[HTML]{EFEFEF}65.34                                  & \cellcolor[HTML]{EFEFEF}59.18                                  & \cellcolor[HTML]{EFEFEF}37.12                                  & \cellcolor[HTML]{EFEFEF}75.57                                  & \cellcolor[HTML]{EFEFEF}59.71                                  \\
                       & LLM Pruner (+ fine-tuning)~\cite{Ma:2023} (NeurIPS, 2023) & 65.11          & 68.11          & 63.43          & 37.88          & 76.44          & 62.19          \\
                       & \cellcolor[HTML]{EFEFEF}Shortened LLaMA~\cite{Kim:2024} (ICLR, 2024) (\ddag)                           & \cellcolor[HTML]{EFEFEF}\textbf{68.82}                         & \cellcolor[HTML]{EFEFEF}69.82                                  & \cellcolor[HTML]{EFEFEF}64.06                                  & \cellcolor[HTML]{EFEFEF}39.93                                  & \cellcolor[HTML]{EFEFEF}74.65                                  & \cellcolor[HTML]{EFEFEF}63.46                                  \\
                       & DISP-LLM~\cite{Gao:2024disp} (NeurIPS, 2024)                  & 64.72          & 68.39          & 64.81          & 37.12          & 76.66          & 62.34          \\
                       & \cellcolor[HTML]{EFEFEF}PruneNet~\cite{YOPO:2025} (ICLR, 2025)                  & \cellcolor[HTML]{EFEFEF}62.12         & \cellcolor[HTML]{EFEFEF}65.40          & \cellcolor[HTML]{EFEFEF}64.65          & \cellcolor[HTML]{EFEFEF}36.52         & \cellcolor[HTML]{EFEFEF}75.51         & \cellcolor[HTML]{EFEFEF}60.82          \\
\multirow{-6}{*}{20\%} & \textbf{AMP (Ours)}                                              & 63.93                                  & \textbf{70.34}                         & \textbf{69.82}                         & \textbf{41.38}                         & \textbf{77.15}                         & \textbf{64.52}                         \\ \hhline{=|=|=====|=}
0\%                    & LLaMA-2 7B                                      & 69.14                                  & 75.99                                  & 74.58                                  & 46.15                                  & 79.11                                  & 68.99                                  \\ \hline
                       & \cellcolor[HTML]{EFEFEF}SliceGPT~\cite{Ashkboos:2024}  (ICLR, 2024)                                         & \cellcolor[HTML]{EFEFEF}61.33                                  & \cellcolor[HTML]{EFEFEF}49.62                                  & \cellcolor[HTML]{EFEFEF}51.77                                  & \cellcolor[HTML]{EFEFEF}31.23                                  & \cellcolor[HTML]{EFEFEF}63.55                                  & \cellcolor[HTML]{EFEFEF}51.50                                  \\
                       & LLM Surgeon~\cite{Ouderaa:2024} (ICLR, 2024)                                       & 61.09                                  & 60.72                                  & 63.09                                  & 36.69                                  & 73.56                                  & 59.03                                  \\
                       & \cellcolor[HTML]{EFEFEF}Shortened LLaMA~\cite{Kim:2024} (ICLR, 2024) (\ddag)    & \cellcolor[HTML]{EFEFEF}61.09          & \cellcolor[HTML]{EFEFEF}54.97          & \cellcolor[HTML]{EFEFEF}45.96          & \cellcolor[HTML]{EFEFEF}34.81          & \cellcolor[HTML]{EFEFEF}60.99          & \cellcolor[HTML]{EFEFEF}51.56          \\
                       & DISP-LLM~\cite{Gao:2024disp} (NeurIPS, 2024)                                           & \textbf{63.93}                         & 62.87                                  & 60.1                                   & 37.03                                  & 73.72                                  & 59.53                                  \\
                       & \cellcolor[HTML]{EFEFEF}PruneNet~\cite{YOPO:2025} (ICLR, 2025)                  & \cellcolor[HTML]{EFEFEF}61.09         & \cellcolor[HTML]{EFEFEF}58.30          & \cellcolor[HTML]{EFEFEF}53.20          & \cellcolor[HTML]{EFEFEF}32.94         & \cellcolor[HTML]{EFEFEF}71.11         & \cellcolor[HTML]{EFEFEF}55.33          \\
                       & PruneNet (+ fine-tuning)~\cite{YOPO:2025} (ICLR, 2025)                  & 62.90         & 63.21          & 53.37          & 33.70          & 72.20         & 57.08          \\
\multirow{-7}{*}{30\%} & \cellcolor[HTML]{EFEFEF}\textbf{AMP (Ours)}                        & \cellcolor[HTML]{EFEFEF}61.25          & \cellcolor[HTML]{EFEFEF}\textbf{65.47} & \cellcolor[HTML]{EFEFEF}\textbf{64.31} & \cellcolor[HTML]{EFEFEF}\textbf{39.85} & \cellcolor[HTML]{EFEFEF}\textbf{74.21} & \cellcolor[HTML]{EFEFEF}\textbf{61.02} \\ \hhline{=|=|=====|=}
0\%                    & Phi-1.5 (1.3B)                                    & 72.77                                  & 62.58                                  & 73.11                                  & 48.04                                  & 75.63                                  & 66.43                                  \\ \hline
                       & \cellcolor[HTML]{EFEFEF}SliceGPT~\cite{Ashkboos:2024} (ICLR, 2024)                                          & \cellcolor[HTML]{EFEFEF}\textbf{64.96}                         & \cellcolor[HTML]{EFEFEF}42.54                                  & \cellcolor[HTML]{EFEFEF}53.66                                  & \cellcolor[HTML]{EFEFEF}31.91                                  & \cellcolor[HTML]{EFEFEF}65.45                                  & \cellcolor[HTML]{EFEFEF}51.70                                  \\
                       & \textbf{DISP-LLM~\cite{Gao:2024disp} (NeurIPS, 2024)}                   & 61.48          & \textbf{47.97} & 57.66          & 33.01          & \textbf{71.08} & \textbf{54.24} \\ 
\multirow{-3}{*}{30\%} & \cellcolor[HTML]{EFEFEF}AMP (Ours)                                               & \cellcolor[HTML]{EFEFEF}58.33                                  & \cellcolor[HTML]{EFEFEF}45.74                                  & \cellcolor[HTML]{EFEFEF}\textbf{58.38}                         & \cellcolor[HTML]{EFEFEF}\textbf{35.32}                         & \cellcolor[HTML]{EFEFEF}69.53                                  & \cellcolor[HTML]{EFEFEF}53.46                                  \\ \hhline{=|=|=====|=}
0\%                    & Phi-2 (2.7B)                                      & 75.61                                  & 73.86                                  & 78.24                                  & 54.01                                  & 79.11                                  & 72.17                                  \\ \hline
                       & \cellcolor[HTML]{EFEFEF}SliceGPT~\cite{Ashkboos:2024} (ICLR, 2024)                                          & \cellcolor[HTML]{EFEFEF}63.14                                  & \cellcolor[HTML]{EFEFEF}47.56                                  & \cellcolor[HTML]{EFEFEF}53.03                                  & \cellcolor[HTML]{EFEFEF}30.29                                  & \cellcolor[HTML]{EFEFEF}65.94                                  & \cellcolor[HTML]{EFEFEF}51.99                                  \\
                       & DISP-LLM~\cite{Gao:2024disp}  (NeurIPS, 2024)                  & 65.19 & 54.43 & 63.59 & 38.48 & \textbf{73.34} & 59.01 \\
                       & \cellcolor[HTML]{EFEFEF}\textbf{PruneNet~\cite{YOPO:2025} (ICLR, 2025)}                  & \cellcolor[HTML]{EFEFEF}\textbf{67.48}         & \cellcolor[HTML]{EFEFEF}56.80          & \cellcolor[HTML]{EFEFEF}\textbf{67.55}          & \cellcolor[HTML]{EFEFEF}\textbf{40.61}         & \cellcolor[HTML]{EFEFEF}72.80         & \cellcolor[HTML]{EFEFEF}\textbf{61.05}         \\
                       & PruneNet (+ fine-tuning)~\cite{YOPO:2025} (ICLR, 2025)                  & 63.93         & \textbf{58.18}          & 61.78          & 37.80          & 71.49         & 58.34          \\
\multirow{-5}{*}{30\%} & \cellcolor[HTML]{EFEFEF}AMP (Ours)                                                & \cellcolor[HTML]{EFEFEF}57.22                                  & \cellcolor[HTML]{EFEFEF}45.89                                  & \cellcolor[HTML]{EFEFEF}61.03                                  & \cellcolor[HTML]{EFEFEF}36.69                                  & \cellcolor[HTML]{EFEFEF}70.95                                  & \cellcolor[HTML]{EFEFEF}54.36                                  \\ \hhline{=|=|=====|=}
\end{tabular}
\caption*{(\ddag) For the LLaMA 7B model, the authors report an 18\% pruning ratio. For a fair comparison, we reproduce the results for a closer pruning ratio (21.02\% -- equal to ours) using the procedures of the paper. In addition, since the authors do not present results for the LLaMA-2 7B model, we run their method on our own.}
\label{tab:state_of_the_art}
\end{table*}


Overall, our method achieves state-of-the-art accuracy across multiple tasks within the LLaMA family. Specifically, for the LLaMA 7B model, we prune 20\% of its parameters and achieve the best results compared to other pruning methods. For the LLaMA-2 7B model, we reduce parameters by 30\%, with only an 11.55\%  drop in average performance relative to the original model. These results confirm that projecting the data onto the weights helps identify the least important structures, making it possible to outperform far more computationally expensive methods, such as SliceGPT~\cite{Ashkboos:2024} and DISP-LLM~\cite{Gao:2024disp}, using a more efficient approach.

Although AMP delivers comparable performance for the LLaMA family, Phi models exhibit a more pronounced  drop. Since smaller models generally have less capacity than larger ones, they may be more sensitive to pruning, leading to greater performance declines~\cite{Frantar:2023}.

\begin{figure}[!bt]
  \centering
  \includegraphics[width=\linewidth]{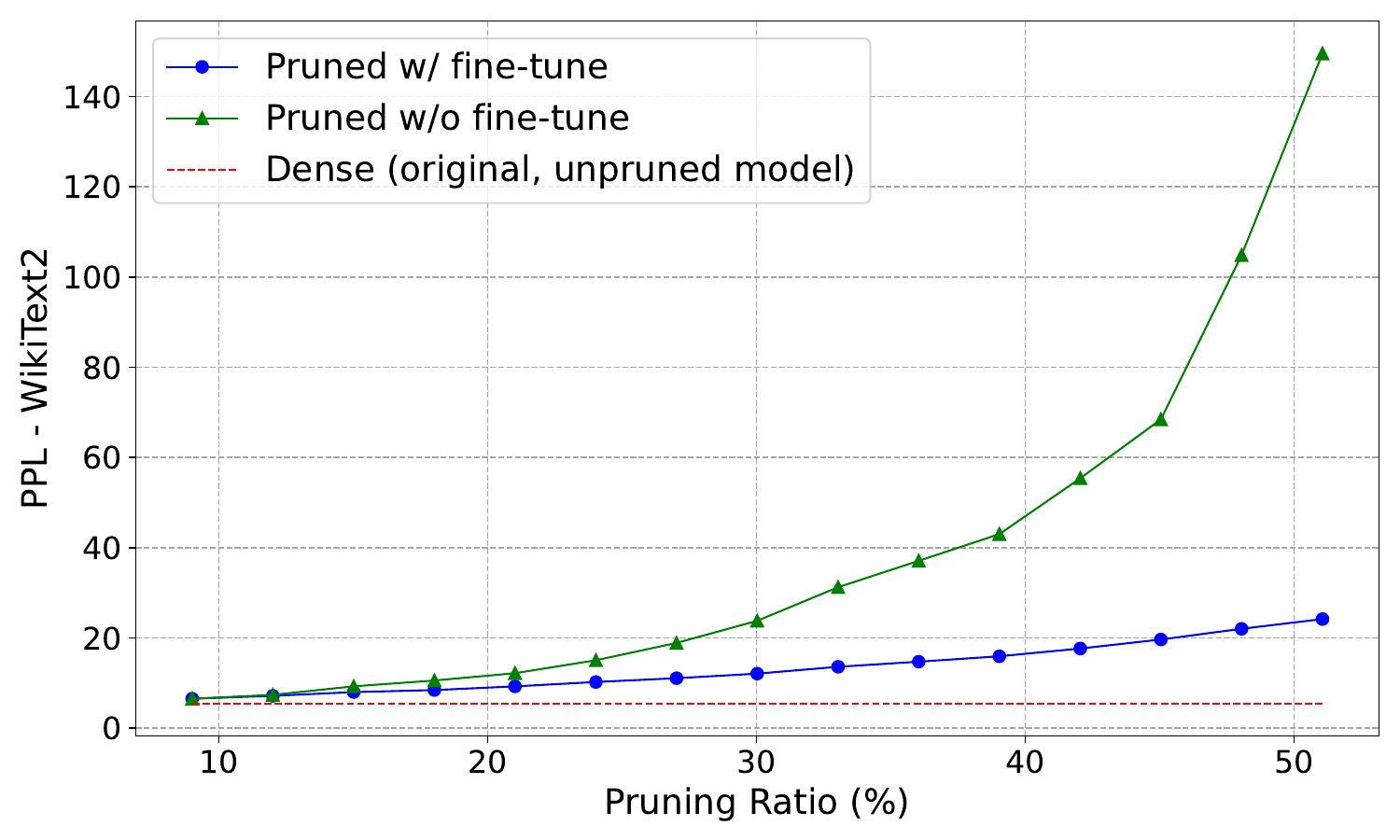}
  \caption{Pruning impact of AMP on WikiText2 perplexity for LLaMA-2 7B. The red dashed horizontal line indicates the performance of the original model, while the blue and green lines represent the performance of the pruned model with and without fine-tuning, respectively.}
  \label{fig:perplexity_pruning}
\end{figure}

Figure~\ref{fig:perplexity_pruning} shows PPL results on the WikiText-2 dataset. As lower PPL indicates better performance, the figure reveals an increasing deterioration in LLaMA-2 7B as the pruning ratio rises. This highlights the critical role of post-training in restoring model performance. In particular, the curve with fine-tuning yields significantly lower PPL values compared to the no-fine-tuning counterpart, and this gap even widens at higher pruning ratios. At 51.06\% pruning ratio, the no-fine-tuning approach exhibits a PPL 6.18 times higher than the former.

As a final note, Kim et al.~\cite{Kim:2024} reported abnormally high PPL values in certain instances. In contrast, our AMP criterion exhibits consistently defined and reasonable PPL values across all compression rates. These findings reinforce its effectiveness compared to top-performance methods.
%

\noindent
\subsection{Inference Speedup}

Even though the number of parameters offers theoretical insights about performance, it may fail to fully capture the model efficiency when used as a standalone metric~\cite{Dehghani:2022}. From this perspective, we include a latency assessment to obtain a more accurate view of the efficiency offered by our method.





We consider LLaMA 7B and LLaMA-2 7B models for inference speedup evaluation. The mean latency results are 2.90s and 2.79s for the unpruned LLaMA 7B and LLaMA-2 7B, respectively.  After pruning~\footnote{For both models, we consider a pruning ratio of 21.02\%.}, the mean latency across runs for LLaMA 7B is 2.31s, representing a speedup of 1.25$\times$, while for LLaMA-2 7B, the mean latency is 2.34s, corresponding to a speedup of 1.19$\times$.
These results demonstrate that our method achieves practical inference speedup and is competitive with other pruning methods. In particular, we surpass methods such as Shortened LLaMA~\cite{Kim:2024} and rivalize with semi-structured approaches like Wanda~\cite{Sun:2024}. The former achieves a $1.13\times$ speedup by pruning entire layers and reducing model depth and the latter delivers an end-to-end inference speedup of 1.24$\times$ on LLaMA 7B  by leveraging sparse tensor acceleration, thus requiring specialized hardware.

%



\noindent
\subsection{Coherence Check of Proposed AMP}


 To confirm the effectiveness of our method in selecting the least important structures for removal, we propose an experiment to check its coherence. Instead of removing the least important structures to achieve a target pruning ratio, we remove the \emph{most important} defined by our criterion. Intuitively, we expect model collapse as an outcome of this reverse pruning, showing results much worse than those of the original tests. Furthermore, we also report the results for random pruning, as previous works suggest its effectiveness~\cite{Wang:2022,Li:2022}. In this experiment, we use the LLaMA-2 7B model at three compression rates across different benchmarks.


\begin{table}[b]
\caption{Coherence check of the AMP method. We highlight in bold the best average accuracy across tasks for each pruning ratio. Based on the results, we confirm that our method effectively identifies and removes structures of lower importance.}
\smallskip\noindent
\resizebox{\linewidth}{!}{%
\begin{tabular}{c|c|ccccc|c}
\hline
Pruning Ratio             & Method                         & WinoGrande                    & HellaSwag                     & ARC-e                         & ARC-c                         & PIQA                          & Avg                           \\ \hline
0\%                       & LLaMA 2 - 7B                   & 69.06                         & 76.02                         & 74.54                         & 46.16                         & 79.05                         & 68.97                         \\ \hline
                          & \textbf{AMP}                       & 65.98                         & 75.02                         & 72.81                         & 47.10                         & 77.58                         & \textbf{67.70}                \\
                          & \cellcolor[HTML]{EFEFEF}Random & \cellcolor[HTML]{EFEFEF}59.27 & \cellcolor[HTML]{EFEFEF}69.63 & \cellcolor[HTML]{EFEFEF}65.74 & \cellcolor[HTML]{EFEFEF}41.13 & \cellcolor[HTML]{EFEFEF}76.17 & \cellcolor[HTML]{EFEFEF}62.39 \\
\multirow{-3}{*}{9.01\%}  & Reversed                       & 49.25                         & 26.55                         & 26.85                         & 27.47                         & 49.89                         & 36.00                         \\ \hline
                          & \textbf{AMP}                       & 61.56                         & 69.22                         & 68.18                         & 42.06                         & 76.39                         & \textbf{63.48}                \\
                          & \cellcolor[HTML]{EFEFEF}Random & \cellcolor[HTML]{EFEFEF}58.72 & \cellcolor[HTML]{EFEFEF}62.83 & \cellcolor[HTML]{EFEFEF}54.21 & \cellcolor[HTML]{EFEFEF}35.92 & \cellcolor[HTML]{EFEFEF}73.18 & \cellcolor[HTML]{EFEFEF}56.97 \\
\multirow{-3}{*}{21.02\%} & Reversed                       & 47.91                         & 26.32                         & 26.89                         & 27.05                         & 50.49                         & 35.73                         \\ \hline
                          & \textbf{AMP}                       & 61.25                         & 65.47                         & 64.31                         & 39.85                         & 74.21                         & \textbf{61.02}                \\
                          & \cellcolor[HTML]{EFEFEF}Random & \cellcolor[HTML]{EFEFEF}55.25 & \cellcolor[HTML]{EFEFEF}53.94 & \cellcolor[HTML]{EFEFEF}50.55 & \cellcolor[HTML]{EFEFEF}30.89 & \cellcolor[HTML]{EFEFEF}68.82 & \cellcolor[HTML]{EFEFEF}51.89 \\
\multirow{-3}{*}{30.03\%} & Reversed                       & 48.22                         & 26.28                         & 25.51                         & 27.56                         & 49.67                         & 35.45                         \\ \hline
\end{tabular}
}

\label{tab:reversed_results}
\end{table}

Table \ref{tab:reversed_results} shows the results and supports that the proposed metric effectively identifies and ranks less critical structures for pruning. It turns out that when the method is reversed -- meaning the most important components are pruned -- the results degrade significantly, with an accuracy drop of more than 25 percentage points across all pruning ratios compared to the original ranking. Interestingly, random pruning, although less effective than our original method, still achieves better performance than the reversed-importance approach. In summary, this contrast reinforces that AMP successfully differentiates between essential and non-essential structures, validating the strategy of selecting lower-importance elements for pruning.

We believe this experiment is essential for validating any pruning criteria, as it is simple and mitigates the influence of hyperparameter variations and other confounding factors.

\noindent
\subsection{The Role of the MHA and MLP Components in AMP}
%
To evaluate the individual contributions of the attention head and MLP components within AMP, in this experiment, we conduct an ablation study by separately pruning attention heads and MLP neurons, and compare the results against applying the full AMP criterion that eliminates both components. We follow the same setup as in Subsection~\ref{exp:main_experiment}.


Table \ref{tab:ablation} exhibits the results for a 30\% compression rate on LLaMA-2 7B. According to this table, we observe that pruning only attention heads severely impairs predictive performance, reducing the average accuracy by 23.39 percentage points (pp) compared to pruning both attention heads and MLPs. It turns out that, for all the models we analyze, MHA accounts for approximately 33\% of the total parameters in a layer. Thus, achieving a compression rate of 30\% by pruning only attention heads would require removing nearly all of them — for LLaMA-2 7B, 30 out of the 32 heads in each layer — leading to large performance drops. On the other hand, pruning MLP neurons, which constitute approximately $67\%$ of the parameters within a layer, provides a more effective reduction in model size compared to pruning attention heads alone. However, our experiment shows that the removal of only MLP neurons also leads to a decrease in performance, with an average accuracy drop of $4.09$ pp compared to pruning both components together.

Overall, by removing all components together, AMP ensures that no component is removed prematurely, allowing higher compression rates while maintaining predictive ability. We plan to further investigate the use of different pruning ratios for attention heads and MLP neurons in future work.




\begin{table}[!tb]
\caption{Influence of removing isolated components and all at once, using LLaMA-2 7B with a 30\% compression rate.  The results emphasize that AMP, which combines the pruning of Heads and MLP neurons (first row), is substantially better than the removal of only one of them.}

\smallskip\noindent
\resizebox{\linewidth}{!}{%
\begin{tabular}{c|ccccc|c}
\hline
Method         & WinoGrande    & HellaSwag    & ARC-e        & ARC-c        & PIQA         & Avg           \\ \hline
\textbf{AMP w/ MHA + MLP Components}       & 61.25       & 65.47       & 64.31       & 39.85       & 74.21       & \textbf{61.02} \\
\cellcolor[HTML]{EFEFEF} AMP w/ MLP Component & \cellcolor[HTML]{EFEFEF}57.14 & \cellcolor[HTML]{EFEFEF}60.64 & \cellcolor[HTML]{EFEFEF}59.81 & \cellcolor[HTML]{EFEFEF}34.64 & \cellcolor[HTML]{EFEFEF}72.42 & \cellcolor[HTML]{EFEFEF}56.93 \\
AMP w/ MHA Component     & 49.49       & 28.13       & 30.60       & 24.23       & 55.71       & 37.63       \\ \hline
\end{tabular}
}
\label{tab:ablation}
\end{table}

\noindent
\subsection{Green AI and Financial Costs}
\label{sub:greenai}



The concept of Green AI has gained attention among researchers, emphasizing the necessity of lowering the computational demands associated with developing and deploying AI models, aiming to minimize environmental impact \cite{Schwartz:2020,Faiz:2024,anonymous2025holistically}. In particular, LLMs are well-known for their high computational costs, both in training and deployment. Hence, they require increasingly large computational resources due to the scaling laws governing their development~\cite{Hoffmann:2022}.

In line with Green AI principles, our AMP method achieves up to 1.25$\times$ inference speedup, representing a 17.39\% direct decrease in operational costs. Furthermore, we achieve a reduction of approximately 19.97\% in carbon emissions\footnote{For reproducibility, we estimate these values using the Machine Learning Impact calculator~\cite{Lacoste:2019}.}. The complete pruning and post-training processes of AMP take approximately four hours to complete using a GPU RTX 3090, with the pruning process itself consuming only a few minutes. These computational costs are typically negligible when compared to the expenses of running these models in production over extended periods.


\noindent
\subsection{Impact on the Number of Fine-tuning Epochs}


To assess the impact of training duration on recovery, we fine-tune the LLaMA-2 7B model at compression rates of 21.02\% and 30.03\% over 1 to 4 epochs. We employ the same hyperparameters and benchmarks as those in Subsection \ref{exp:main_experiment}.

For a compression rate of 21.02\%, the average results across tasks are 64.12\%, 63.44\%, 63.41\% and 62.56\% for 1, 2, 3, and 4 epochs of fine-tuning, respectively. For a compression ratio of 30.03\%, the results are 60.72\%, 61.02\%, 60.10\% and 59.18\%. From these results, we observe that using more than one epoch of fine-tuning does not significantly improve model performance. Therefore, we believe a more promising approach would be to train on much larger datasets instead of using more epochs, since Alpaca contains only 52k samples.
However, we opt to use two epochs to create a fair comparison with other works, such as LLM-Pruner~\cite{Ma:2023}.

%% file: Sections/Discussion.tex
\section{Conclusions}\label{sec:discussion}

We introduce AMP, a novel structured pruning method designed to efficiently identify and eliminate less critical components (attention heads and MLP neurons) from Large Language Models (LLMs), producing highly optimized models. Moreover, our coherence check confirms that AMP correctly ranks the importance of each component. Particularly, when our method removes components ranked as more critical (those it should preserve), the model performance drastically decreases. In addition, our method is computationally efficient, requiring only a few minutes on a consumer-grade GPU (NVIDIA RTX 3090) to identify the components for removal.

Through extensive experiments, we show that the proposed method outperforms state-of-the-art structured pruning approaches. Specifically, AMP surpasses existing techniques by a large margin, achieving mean accuracy improvements of up to 4.81 pp for LLaMA 7B and 9.52 pp for LLaMA-2 7B. Furthermore, perplexity evaluation on WikiText2 corroborates the results by demonstrating the consistency of AMP across many compression rates. Notably, AMP achieves up to 1.25× inference speedup  without requiring specialized hardware, further enhancing efficiency and sustainability. Finally, apart from these benefits, our method aligns with Green AI principles, emphasizing the role of AMP in reducing the computational costs of LLMs. 

Last but not least, we present evidence of the significance of the post-training stage in restoring model performance after pruning, corroborating previous work~\cite{Xia:2024}. In this context, we hypothesize that extending post-training on larger datasets would yield even greater performance, a direction we plan to explore in future work.

%% file: Sections/Acknowledgment.tex
\section{Acknowledgments}

The authors would like to thank Instituto de Ciência e Tecnologia Itaú (ICTi) and Programa de Bolsas Itaú (PBI). This study was financed, in part, by the São Paulo Research Foundation (FAPESP), Brasil. Process Number \#2023/11163-0. This study was financed in part by the Coordenação de Aperfeiçoamento de Pessoal de Nível Superior – Brasil (CAPES) – Finance Code 001. The authors would like to thank grant \#402734/2023-8, National Council for Scientific and Technological Development (CNPq). Artur Jordao Lima Correia would like to thank Edital Programa de Apoio a Novos Docentes 2023. Processo USP nº: 22.1.09345.01.2. Anna H. Reali Costa would like to thank grant \#312360/2023-1 CNPq. 
This work was partially supported by the Instituto Nacional de Ciência e Tecnologia em Inteligência Artificial Responsável para Linguística Computacional, Tratamento e Disseminação de Informação (INCT-TILDIAR - CNPq grant \#408490/2024-1).